\documentclass[acmlarge]{acmart}

\setcopyright{none}

\makeatletter
\def\@mkbibcitation{}
\def\@setauthorsaddresses{}
\def\@copyrightpermission{}
\def\@copyrightowner{}

\makeatother
\fancypagestyle{firstpagestyle}{
\fancyhead{}
\fancyfoot{}
}
\fancypagestyle{mypagestyle}{
\fancyhead[LE,RO]{\thepage}
\fancyhead[LO,RE]{}
\fancyfoot[LE,LO,RE,RO]{}
}
\usepackage[noend]{algpseudocode} 
\usepackage[ruled]{algorithm}
\usepackage{algorithmicx}

\floatname{algorithm}{\small ALGORITHM}
\usepackage{etoolbox}
\AtBeginEnvironment{algorithmic}{\small}

\usepackage{xspace}

\usepackage{array}
\usepackage{tabularx}
\usepackage{booktabs}
\newcolumntype{R}[1]{>{\raggedleft\arraybackslash}p{#1}}
\newcolumntype{L}[1]{>{\raggedright\arraybackslash}p{#1}}
\newcolumntype{C}[1]{>{\centering\arraybackslash}p{#1}}
\newcolumntype{A}{>{\raggedright\arraybackslash}X}
\newcolumntype{B}[1]{>{\raggedright\arraybackslash}b{#1}}

\usepackage{enumitem}

\usepackage[export]{adjustbox}
\usepackage{subcaption}

\usepackage{soul}
\definecolor{myhlcolor}{HTML}{DDDDDD}
\sethlcolor{myhlcolor}

\begin{document}
\title[Prediction Factory]{Prediction Factory: automated development and collaborative evaluation of predictive models}


\begin{abstract}

In this paper, we present a data science automation system called Prediction Factory. The system uses several key automation algorithms to enable data scientists to rapidly develop predictive models and share them with domain experts. To assess the system's impact, we implemented 3 different interfaces for creating predictive modeling projects: baseline automation, full automation, and optional automation. With a dataset of online grocery shopper behaviors, we divided data scientists among the interfaces to specify prediction problems, learn and evaluate models, and write a report for domain experts to judge whether or not to fund to continue working on. In total, 22 data scientists created 94 reports  that were judged 296 times by 26 experts. In a head-to-head trial, reports generated utilizing full data science automation interface reports were funded 57.5\% of the time, while the ones that used baseline automation were only funded 42.5\% of the time. An intermediate interface which supports optional automation generated reports were funded 58.6\% more often compared to the baseline. Full automation and optional automation reports were funded about equally when put head-to-head. These results demonstrate that Prediction Factory has implemented a critical amount of automation to augment the role of data scientists and improve business outcomes. 

\end{abstract}

%
%

%
%


\maketitle



\section{Introduction} 
A growing need for data science experts and lack of their availability has highlighted the benefits of automating many of the functions those experts perform. Each year, the machine learning and data mining communities further expand the possibilities for such automation. First came the ability to automatically identify the best hyperparameters for a machine learning model when presented with a data set consisting of features and labels \cite{thornton2013auto,snoek2012practical}. The subfield known as \textsc{automl}, next focused on automated methods for simultaneous model selection and hyperparameter tuning. This automation ultimately extended to tuning the hyperparameters for an entire machine learning pipeline, including transformations such as PCA, albeit still starting with features and labels \cite{feurer2015efficient}. Recognizing that ideating, writing scripts and extracting features for a given prediction problem could consume up to 90\% of available time, multiple teams also developed automated methods for feature engineering \cite{tran2016feature, kanter2015deep,katz2016explorekit}. The existence of tools that can automate these two steps can assist data scientists in two ways. First, not having to manually tweak hyperparameters and test models or engineer features frees up their time. Second, this automation may help them discover a highly accurate model. 

Notably, in order to begin using these tools, a user must have a well-defined prediction/machine learning task, along with task-specific, pre-processed and prepared data. Even after a prediction problem definition is chosen, it is often necessary to set prediction-specific parameters - how far ahead does one want to predict?  After examining how business questions and other real-world problems are translated into machine learning or predictive modeling tasks, authors in \cite{kanter2016label} discovered that defining these tasks takes a significant amount of time.

We found that while the two existing automation techniques are \textit{necessary}, they are not sufficient. Two more things are required: a structured way of defining a prediction problem and specifying its parameters, and an automatic way of processing the data once the problem is defined. These two steps called ``prediction engineering'' are addressed through another automation system developed by authors \cite{kanter2016label}.


With these three automation systems available to us we can generate numerous prediction problems and find solutions for them automatically. Our next challenge is to select one with high impact for business. Currently, the decision of which ``prediction problem'' to solve is made up front, without full knowledge of how accurate the model would be, or whether a problem could be solved using the data at hand. Additionally, choosing between multiple possible solutions is subjective and requires domain expertise. It is not unusual for two or more disparate prediction tasks to both constitute likely solutions for a business problem. For example, to address customer churn, one may attempt to predict likelihood of churn, or may attempt to predict what product to recommend next to best serve the customer.

To enable this, we propose Prediction Factory.  Backed with automated tools, with Prediction Factory, data scientists can create predictive models using automated tools, and write reports communicating their results. Domain experts can then view the reports and assess potential solutions side by side. We imagine an outcome that consists of a curated set of solutions with a compelling business case.

Ultimately, automation should both enable higher throughput by allowing more problems to be solved using machine learning and data science and broaden participation by lowering the barrier to entry, and thus extending the pool of potential users beyond highly-trained experts. Now that these automated systems are readily available, in this paper we systematically examine whether they achieve the stated goals.

\noindent \textbf{Our major contributions include:}
\begin{itemize}
\item Prediction Factory: a first interactive system for enabling data science through automation.
\item In developing prediction factory and assessing it \textit{via} experiments involving users, we present the first systematic quantitative evaluation of (to the best of our knowledge):
\begin{itemize}
    \item[--] the impact data science automation; 
    \item[--] and what parts of automation are critical for improving business outcomes. 
\end{itemize}
\end{itemize}

The paper is organized as follows: Section~\ref{automation} presents the three automation tasks to achieve full data science or machine learning automation. Section~\ref{related} presents the related work.  Section~\ref{prediction_factory} presents the design of our prediction factory system. Section~\ref{exp_interfaces} presents the experimental interfaces we designed. Section~\ref{experiments} presents the experiments we conducted. Section~\ref{results} presents the results following by discussion and conclusions in sections~\ref{discussion} and ~\ref{conclusion} respectively. 

\begin{table*}[!t]
\begin{tabularx}{\linewidth}{@{}p{3.2cm}X@{}}
\includegraphics[width=3cm,valign=t]{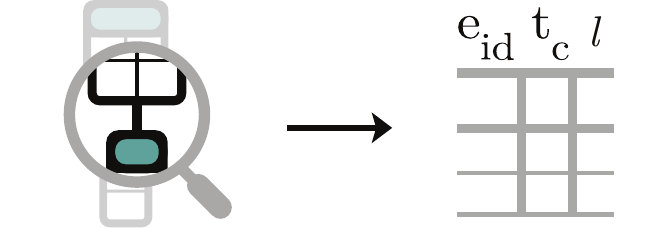} & \textbf{1.\ Prediction engineering - \textsc{autope}}\par
Given $f(.)$, and $D$ come up with labeled training examples (past occurrences of the outcome) as a list of three tuples $<e_{id}, t_c, l>$ where $t_c$ is the latest time made available to the feature engineering and model engineering steps.
\\[1em]
\includegraphics[width=3cm,valign=t]{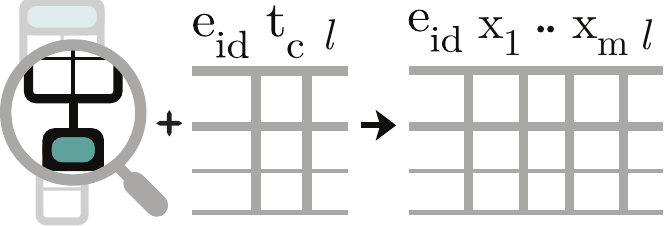} & \textbf{2.\ Feature engineering - \textsc{autofe}}\par
Given $D$ and $<e_{id}, t_c, l>$ generate features $<x_{1\ldots m}>$ for each training example using only the data prior to the time point $t_c$ associated with it.  
\\[1em]
\includegraphics[width=3cm,valign=t]{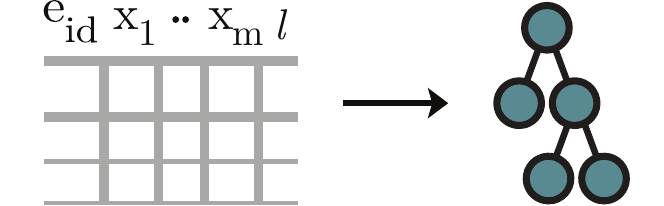} & \textbf{3.\ Model engineering - \textsc{automl}}\par
Given $<x_{1\ldots m}>$ and $<e_{id}, t_c, L>$ generate a model $M (x_{1 \dots m}) \rightarrow l$ that given a feature vector can predict the label ~$l$ 
\end{tabularx}
\medskip
\caption{Three automation tasks and what they should deliver to generate a predictive model from multi entity, multi table, temporal and relational dataset $D$.}\label{table-steps}
\end{table*}
\section{Automation}\label{automation}
Our focus is on multi-entity, multi-table temporal and relational datasets. Commercial, academic and several other entities accumulate these datasets on a day-to-day basis as people interact with a variety of digital, physical and cyber systems. A recent survey by the online data science platform \textsc{Kaggle} found that 65.5\% of data scientists work with this type of dataset (followed by 53.3\% for text). In comparison, only 18.8\% work with image data \footnote{The state of ML and Data Science. Available at \url{https://www.kaggle.com/surveys/2017}}.

A typical machine learning goal when working with this data is to predict an \textit{outcome} ahead of its occurrence. Some popular examples of predictive models that academics, data scientists and others have attempted to build using this type of data source include:
\begin{itemize}
    \item[--] predicting student dropout on massive open online courses;
    \item[--] predicting acute hypotensive events from time series signals collected in an ICU;
    \item[--] predicting customer churn;
    \item[--] several problems hosted on \textsc{Kaggle} also fall under this category.
\end{itemize}

Next we describe the data input for these types of projects, three \textit{engineering} tasks necessary when working with this type of dataset, and what an automation algorithm should deliver. 

\noindent \textbf{Data input}: In these cases, typically a data scientist encounters a data $\mathtt{D}$, which has tables $T_{1 \dots k}$. Each table $T_i$ has columns $C^i_{1 \dots c}$ and a column $C^i_p$ representing a primary key specifying the unique elements of the table. Optionally, a $\mathtt{datetime}$ column, $C^i_t$ represents a $\mathtt{timeindex}$ for the table. The value in $j^{th}$ row $\mathtt{timeindex}$ column represents when data in the $j^{th}$ row was recorded (or became available). A set of foriegn key relations provide the linking between the tables. Each table is referred as an \textit{entity} and each unique row as an instance of that \textit{entity}.

\noindent \textbf{Outcome to predict}: A data scientist also has an outcome that s/he wants to predict. Usually this is expressed programmatically using a function  $f(.)$ that when given $D_{T_{ij}}$ -- an arbitrary slice of data for $j^{th}$ instance of entity (table) $i$ and all its related data -- returns a continuous value or a binary or a categorical value which is the outcome s/he wants to predict. For convenience of explanation, we will consider binary outcomes here.

Given the data $D$ and the outcome function $f(.)$ the three tasks that a data scientist (or a machine learning engineer) performs are:
\begin{itemize}
    \item[--] Prediction engineering: To train a machine learning model to predict an outcome, one must find past occurrences of the outcome as training examples (as well as some examples when the outcome did not happen). For example, to build a predictive model for student dropout, one has to identify all the students that did dropout and the ones that did not from the past courses. This search process is parameterized, enabling the data scientist to control how many examples to extract, the time gap between two examples from the same entity, and several other factors \cite{kanter2016label}. The search results in a list of three tuples: the $id$ of the entity instance, the time point at which the outcome occured and the binary label. The goal for the next task is to use the data prior to the time point and construct a model. 
    \item[--] Feature engineering: The next task for a data scientist is to extract variables (a.k.a features). For each training example in the list above, one must create variables using only the data prior to the timepoint in order to emulate the real-world scenario and prevent label leakage. One example of a feature in the student dropout prediction problem is: ``\textit{how much time did the student spend watching videos before dropping out?}''.  The output of this task is a set of variables $x_{1\dots m}$ for each training example.  
    \item[--] Model engineering: This is the final task, in which the data scientist, given the features $x_{1\dots m}$ and labels $l$, attempts to find the best possible machine learning model, $M$ and fits its parameters. There are numerous model choices including support vector machines, neural networks, and decision trees. Successful model engineering is the ultimate goal of the \textsc{automl} community.
\end{itemize}

In Table~\ref{table-steps} we present the goal of the three automation tasks corresponding to each of these steps and what their output should be. Over the past several years, much of these three steps have been automated (provided that the relevant data and meta-data are available in the format specified above). Open-source tools are available to deliver automation for these three steps. We call the automation tools available for these steps as \textsc{autope}, \textsc{autofe} and \textsc{automl} respectively. 


\section{Related work}\label{related}
Recently, experts in various domains have expressed great interest in the development and application of machine learning algorithms. This has led to a focus on the development of general-purpose abstractions, automation, and methods for addressing the complexities of data. In \cite{wagstaff2012machine}, the authors emphasize that \textit{phrasing a problem as a machine learning task} is a necessary and important step. Works like \cite{warrick2012machine} and \cite{ghassemi2014unfolding} demonstrate the possible impacts when machine learning and domain experts combine to solve major challenges. 

To smooth out these interactions, much focus has been placed on automating various steps, including data cleaning and preparation \cite{stonebraker2013data}, feature or variable engineering \cite{kramer2001propositionalization}, \cite{kanter}, \cite{featuretools}, or providing useful and intuitive interfaces to complex algorithms \cite{hofmann2013rapidminer}. However, to the best of our knowledge, no system exists for generating predictive modeling questions to ask of data. With the aforementioned tools available, we consider it timely to shift our focus towards a system that can automatically generate the questions themselves from this type of data.

Other data science and machine learning researchers have commented on deficiency of between machine learning performance metrics on assessing model value. In 2006, Netflix launched a competition for a better movie recommendation engine and awarded \$1 million prize to the team with the lowest root mean squared error (RMSE). Their conclusion from the best submissions was that "the additional accuracy gains that we measured did not seem to justify the engineering effort needed to bring them into a production environment" \cite{amatriain2013mining}. This supports that idea that measuring the impact of predictive modeling cannot always be reduced to standard performances metrics.
\section{Prediction factory} \label{prediction_factory}
\begin{figure*}[t]
\centering
\includegraphics[width=\textwidth]{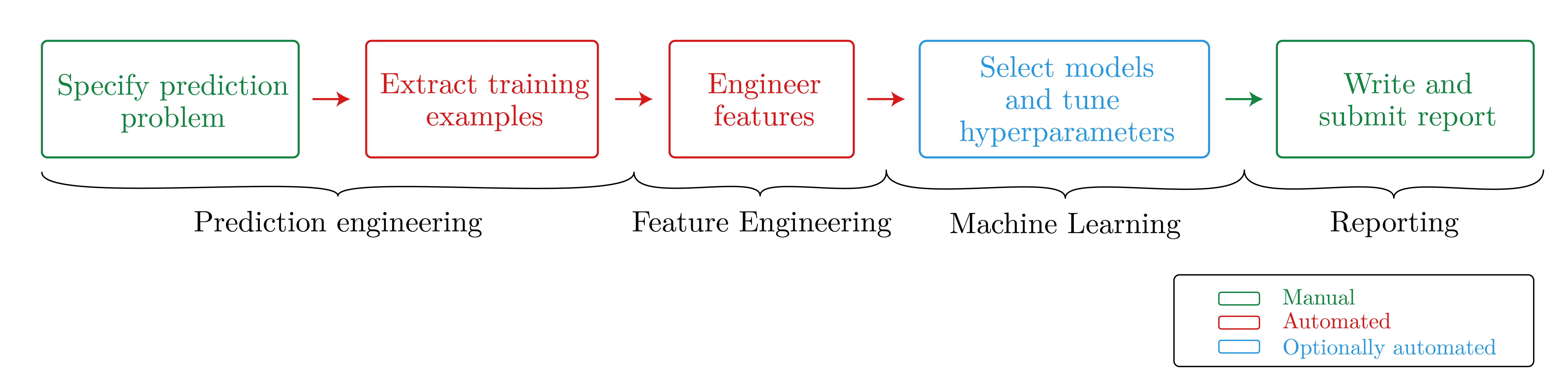}
\caption{ The process for developing a predictive model in Prediction Factory. The first and last steps, "Specify Prediction Problem" and "Write and submit report," require human input, while "Extract Training Examples" and "Engineer features" can be fully automated. "Select models and tune hyperparameters" can be optionally automated}\label{pf-process}
\end{figure*}

Given a dataset, developing a solution for a business problem involves undertaking 5 steps as shown in Figure~\ref{pf-process}. We used the available open source tools for \textsc{automl} and \textsc{autofe} and built our own for \textsc{autope} following the algorithms laid out by authors in \cite{kanter2016label}. With our ability to generate predictive problem definitions, process data and extract training examples, we believe that end-to-end automation of this process is possible.

Automated prediction engineering should enable users to explore more diverse problems, while automated feature engineering, model selection and tuning will help them assess their performance quickly. Users can set parameters for a prediction problem with full knowledge of how they will impact accuracy -- for example, they can ask \textit{"how much accuracy is gained when prediction is attempted 5 days ahead instead of 7 days ahead?"} This has the potential to bring us one step closer to the aforementioned goals for automation -- to increase the number of problems being solved using data, and to provide an easy-to-use interface that will allow business experts to develop, explore and select models. 

However, after enabling this level of automation, two new problems arise:
\begin{itemize}
\item [--] By enumerating outcome functions and their parameters, we are left with an enormous space of problems that we can define. To select a solution, we will have to assess or compare two or more possible solutions. 
\item [--] The enormous problem space makes this assessment and comparison difficult. Since it is possible to generate disparate problem definitions -- for example, predict whether a customer will not return, or predict whether a customer will stop buying from clothing department -- accuracy metrics cannot be used to compare them. A predictive solution should ultimately be assessed in terms of the business metric it will be used to optimize -- cost, efficiency or revenue. At this stage, data scientists and/or business analysts do not have enough knowledge about the uncertainties associated with the system where this model will be deployed, and/or the cost-benefit analysis, to make a properly informed selection.
\end{itemize}
Given that there is no objective way to compare two solutions at this stage, data scientists or domain experts are only able to make subjective judgments based on their past experiences or domain knowledge. And due to the enormous search space, these solutions cannot possibly be judged by a single person or a small group of people. For this reason, we have created what we call "Prediction Factory." With Prediction Factory, data scientists can create predictive solutions using automated tools, and write reports detailing their assessments of predictive solutions. Domain experts can then view the reports and assess potential solutions approaches side by side. This interface lets a large group of people look at the space of possible predictive solutions, and enables a back-and-forth where different experts both create and assess those solutions. We imagine an outcome that consists of a curated set of solutions with a compelling business case. In the next two subsections, we describe our interface in detail. 

\subsection{Creating predictive models}

\begin{figure*}%
    \centering
    \includegraphics[width=0.315\linewidth]{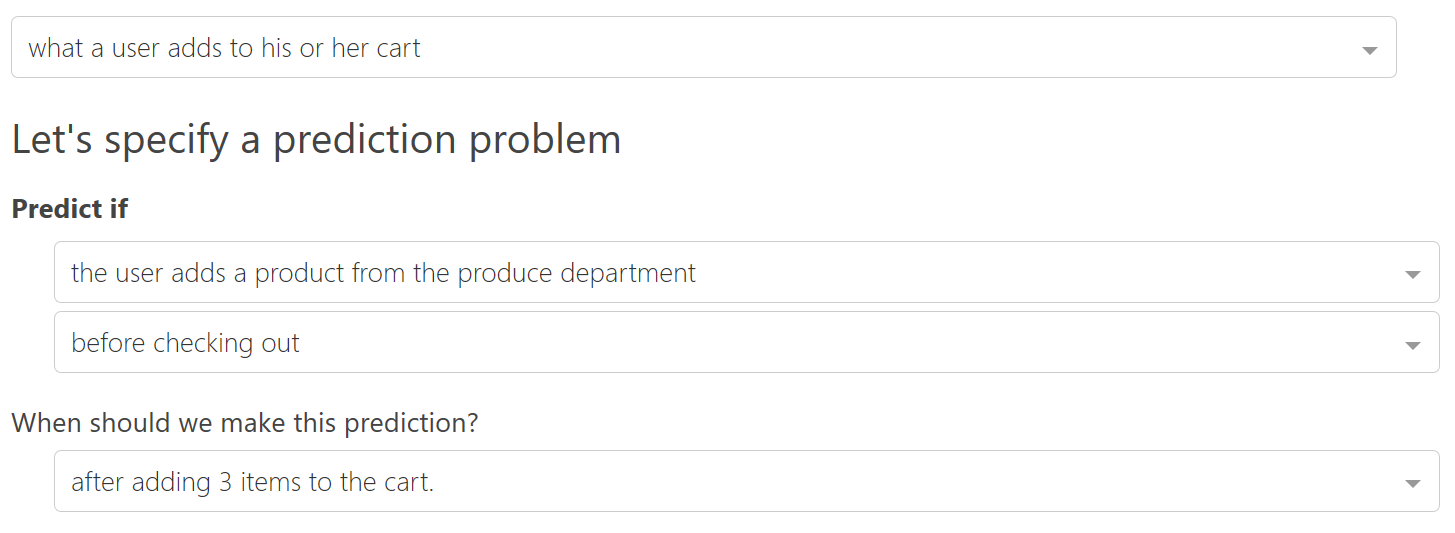}
    \includegraphics[width=0.315\linewidth]{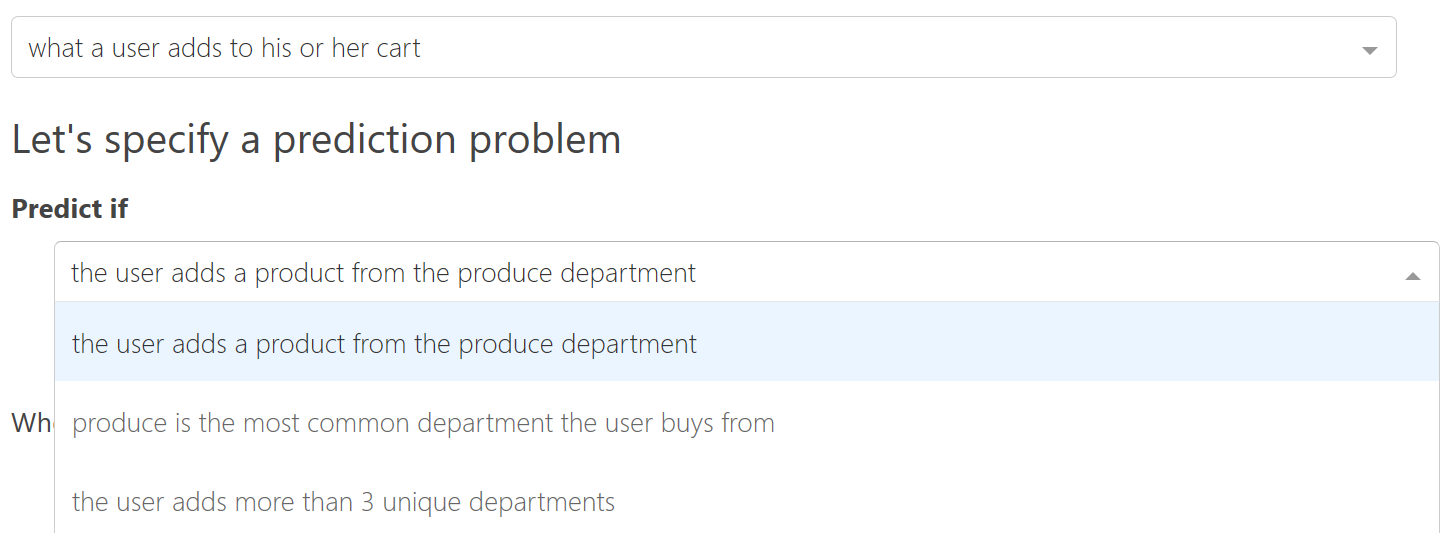}
    \includegraphics[width=0.315\linewidth]{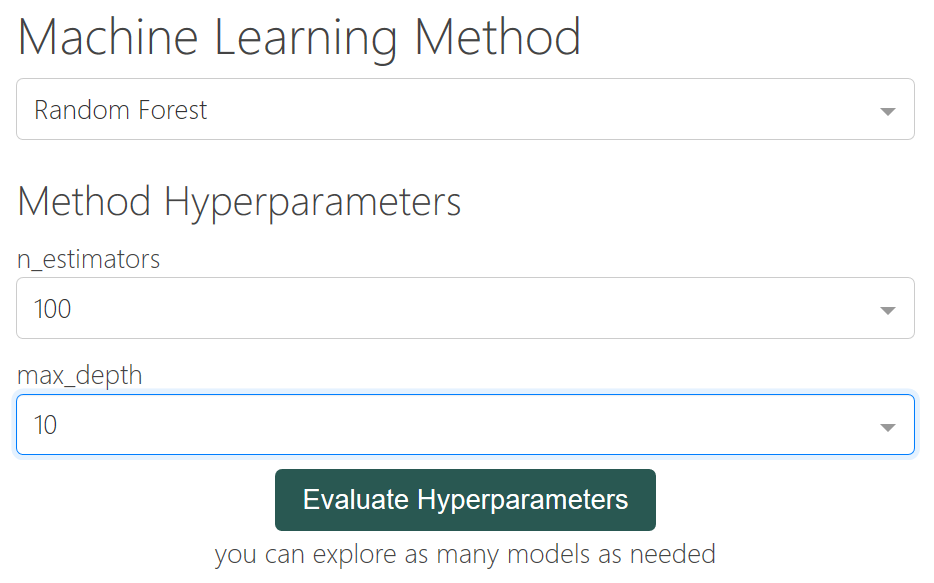}
    
    \caption{From left to right: A human-readable sentence specifying the prediction problem, dropdowns to specify the prediction problem, and the machine learning interface to selecting a model and tune hyperparameters.}%
    \label{fig:define_prediction}%
\end{figure*}

For data scientists who create predictive models, automating both prediction engineering and feature engineering speeds up many steps in the data science process, from defining a prediction problem to generating machine learning-ready data. These scientists can either choose to maintain control over the machine learning model and hyperparameter selection, or let the system make the necessary choices. The process of writing a report and submitting, actions remain subjective and manual. Figure \ref{pf-process} shows the model creation process in Prediction Factory, delineating manual, automated, and optionally automated components. In the "create" interface, the user follows these steps: 

\noindent \textbf{Specify a predictive problem} 
The Prediction problem specification interface includes a button to select the entity on which to make predictions, and for the currently selected entity, a human-readable sentence describing a particular prediction problem. Figure \ref{fig:define_prediction} shows how components of the sentence have dropdown menus that corresponded to problem specific parameters.

Once the user has fully specified a predictive problem definition, automated algorithms scan the data, find the training examples, and create machine-learning-ready features.

\noindent \textbf{Learn a model and evaluate}: Next, the user creates a model. The modeling method and hyperparameter selection interface asks users to select one of many standard machine learning methods. For example, Random Forest (RF), Support Vector Machine (SVM), and Multi-Layer Perceptron (MLP). \footnote{Scikit-learn's implementations for each were used to compute \cite{scikit-learn}.} Each of these methods includes several hyperparameter choices. For a full list of hyperparameter choices for each algorithm, see table \ref{table:hyper_alg_choices}. Once these have been specified, the user can learn and evaluate the model. 

\noindent \textbf{Write a report and submit}: The final piece of the model creation interface shows the results of the selected model in several metrics: F1 (the harmonic mean of precision and recall), ROC-AUC (area under the receiver-operating characteristic curve), and Accuracy. When applicable to the algorithm, it also lists the top 10 features in terms of predictive power. The user then writes his or her own assessment of why this model might be useful, and makes a case for it to be selected when assessed by judges.  Each submission includes free text to explain the relevancy of the model and the modeling decisions that were made.

In summation, the full model creation interface comprises three parts: a part for specifying prediction problems and associated problem-specific parameters, a part for choosing methods and associated hyper-parameters to solve these problems, and a third part that lets users describe their solution using free text before submitting. 

\subsection{Judging or assessing predictive solutions}

Users judge predictive modeling reports through a series of pairwise comparisons. For each comparison, they are given this prompt: ``\textit{The following two predictive modeling projects are presented to you. You have \$1000 at your disposal to fund one of them, which do you choose?}''

\begin{figure}[h]
\centering
\includegraphics[width=0.5\textwidth]{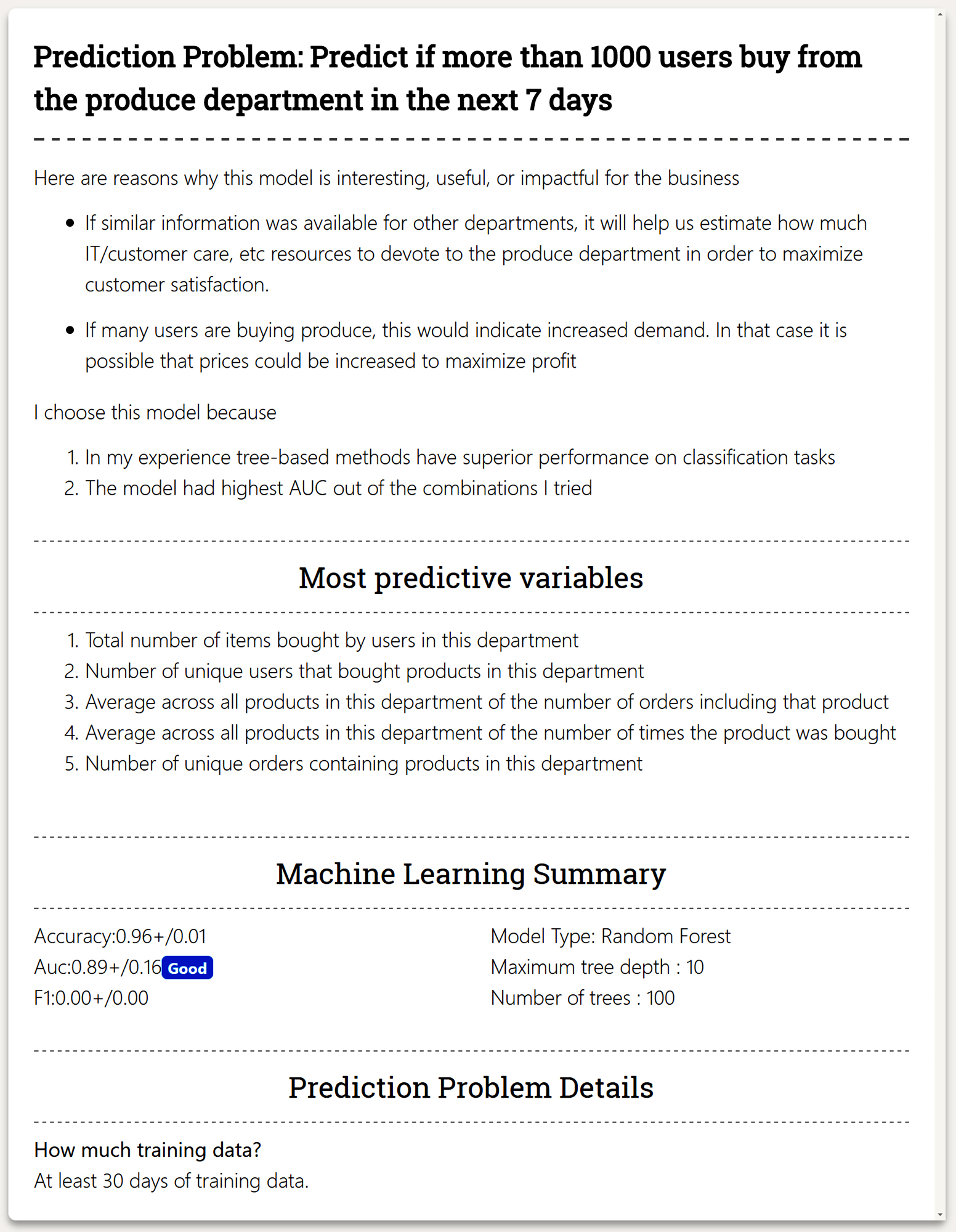}
\caption{An example model report that is displayed in the Model Judging interface}
\label{example-report}
\end{figure}

Modeling reports summarize the predictive models that have been created by other users. Displayed most prominently are the prediction problem as a colloquially written English sentence, and the user-supplied text describing why their project should be funded. Additionally, the report provides standard machine learning performance metrics (Accuracy, AUC, and F1) and ranked feature importance where possible. An example report is shown in Figure  \ref{example-report}.

A judge is able to toggle between two modeling reports at time. To submit a vote, the judge picks one project to fund and provides a 10 - 100 word explanation for the decision. After submitting their vote, a judge may move on to assess other pairs of reports. 

\section{Experimental Modeling Interfaces} \label{exp_interfaces}

By providing the ability to automatically perform prediction and feature engineering, and building off existing work on automatic model selection and tuning, we believe we have created a big leap in data science automation. As with any automated system, we wanted to evaluate its efficacy and its influence on the data scientist's workflow. More specifically, we wanted to know:
\begin{itemize}
    \item [--]Has the system increased the data scientist's throughput? (Throughput is measured by the number of disparate prediction problems solved in a short span of time.)
    \item [--]Has it changed the way the data scientist interacts with the process? In other words, by offering the ability to rapidly develop disparate predictive problems, did we affect how they spend time?
    \item [--]What is the right level of automation? What steps are best automated, and what should be left to manual tuning? 
    \item [--] Has the system broadened participation in the data science process?
\end{itemize}
To address these questions, we created 3 experimental interfaces with different degrees of automation, provided different interfaces to different users, monitored their usage, and collected their submissions. In designing these interfaces, our goal was to provide data scientists and domain experts with a familiar experience similar to that of their day-to-day jobs, with the added advantage of embedded automation tools and an easy-to-use interface for commanding them. 
\subsection{Modeling Interfaces}
\begin{figure}[h]
\centering
\begin{subfigure}[t]{0.3\textwidth}
\includegraphics[width=\textwidth]{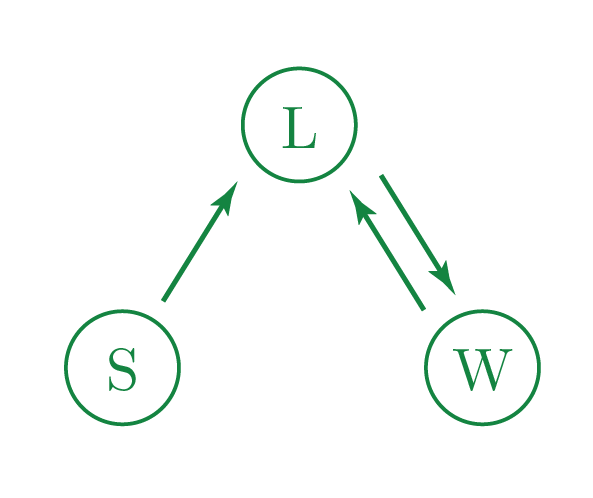}
\caption{}\label{a}
\end{subfigure}
\hspace{1cm}
\begin{subfigure}[t]{0.3\textwidth}
\includegraphics[width=\textwidth]{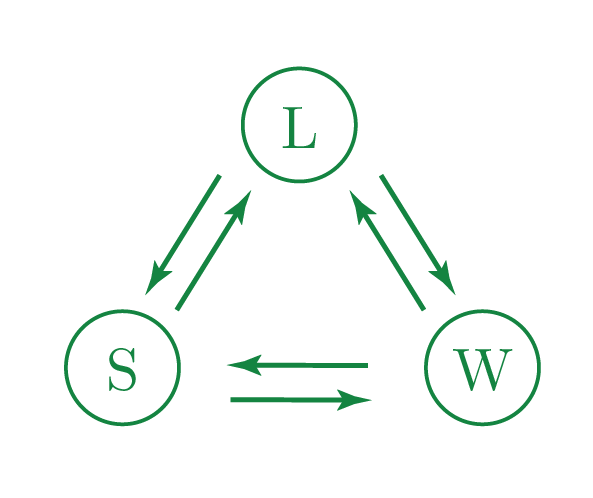}
\caption{}\label{bc}
\end{subfigure}
\caption{(a) Allowed states and transitions for Interface A. Users in this interface must specify a prediction problem (S) before freely learning and evaluating a model (L) and writing and submitting a report (W). (b) Allowed states and transitions for Interfaces B and C. Users in this interface can take advantage of automation while moving between each of the tasks.}
\end{figure}

\begin{description}
 \item \textbf{Interface A: Baseline Automation} This interface allows full control of specifying a problem and learning and evaluating a model, but forces users to finalize  their problem and desired parameters before modeling. Once they specify a problem, prediction engineering and feature engineering is provided to them as an automated service. They then can select the modeling method and hyper parameters and iterate between writing a report and learning a model as many times as they want. Figure~\ref{a} shows how user can switch between the three states in this interface. 
 
 \item[Interface B: Maximum Automation] This interface only allows users to control specifying a problem. Once they specify the prediction problem they can click on a button which automatically extracts training examples, generates features, and learns a  default model with preselected hyperparameters. Users of this interface were allowed to change the prediction problem parameters after viewing modeling results as many times as they want and as they wrote their report.

 \item[Interface C: Optional Automation] This interface gave users full control over both Problem Specification and Algorithm Specification, with the option use a default model with pre-specified hyperparameters. Like interface b, they are allowed to change their prediction problem as many times as they want.
\end{description}

\begin{table}[t]
\centering
\small
    \begin{tabular*}{\linewidth}
        {@{}l@{\extracolsep{\fill}}lll@{}}
        \toprule
         & \textbf{\normalsize Prediction Engineering} & \textbf{\normalsize Feature Engineering} & \textbf{\normalsize Hyperparameter optimization}\\
        \midrule
        Interface A & Yes & For 1 problem & No \\
        Interface B & Yes & Unlimited & Yes \\
        Interface C & Yes & Unlimited & Optional \\
        \bottomrule
    \end{tabular*}
    \caption{Automation capabilities of each interface}
\end{table}

\subsection{User experience}
As users arrived at our interface, they were allowed to choose between the ``\textit{judging}'' and ``\textit{creating}.'' roles. Users were then given directions for their role and presented with a survey that asked them to identify the primary focus of their current job. If they chose ``\textit{creating},'' they were then assigned to one of the three interfaces.

\noindent \textbf{Calculating time on task} \label{timing_details}

Each of the interface groups must progress through the same 3 key tasks. 
\begin{enumerate}
  \item Specifying a prediction problem (labeled S in Figures~\ref{a} and \ref{bc})
  \item Learning a model and evaluating it (L)
  \item Writing a report and submitting it (W)
\end{enumerate}

A user must be performing one, and only one, of these three tasks while using Prediction Factory. We create a timer for each of the three tasks that starts and stops as needed. We also record the user's total time on the page, which is the sum of the time spent on the three tasks. Because users may be using Prediction Factory while doing other tasks on their computer, we pause the timer after 30 seconds of inactivity and restart it when activity resumes.


\section{Experiments}\label{experiments}
We applied Prediction Factory to a dataset of online grocery shopping behavior provided by Instacart, a popular online grocery delivery service. To perform our experiments we collected contributions from dozens of human participants using Prediction Factory through a public web interface. In this section, we explain how we conducted this experiment using Prediction Factory as explained in Section \ref{prediction_factory} and \ref{exp_interfaces}

\subsection{Dataset Details}
We used an anonymized dataset\cite{instacartdataset} containing a sample of over 3 million grocery orders from more than 200,000 Instacart users released on May 3rd, 2017. For each user, the dataset contained between 4 and 100 of their orders, as well as the sequence of products purchased in each order. For each order, the dataset contained the week and hour of day the order was placed, and a relative measure of time since the previous order. The dataset also contained information on the department and aisle each product belonged to. There were a total of 21 departments, 134 aisles, and 49,688 products.

We further processed this dataset to provide an absolute time measured from an arbitrary zero-point for the start of each order and for each cart-add within the order, with cart-adds separated by one millisecond. This arbitrary separation ensured that subsequent orders did not overlap, and preserved the relative sequence of cart-adds within orders. This time-axis is necessary for both our labeling and feature-engineering algorithms.

In order to reduce computation costs, we took a random sampling 5,156 users, which left us with 85,316 orders, and 848,224 total products added to orders. While the predictive performance we saw on this sample may differ from the full dataset, the experiment is primarily concerned with how users interact with and use an automated tool. Furthermore, none of these models were designed for deployment on live data. 


Because grocery shopping is a familiar aspect of most of our lives, the prediction problem space is fairly easy to grasp, even for those not familiar with data science. Furthermore, this particular dataset lent itself to a large space of meaningful problems by virtue of containing multiple entities (namely users, orders, departments, products, and aisles) of which predictions of future information was valuable, as well an interesting time axis to make those predictions over. In section \ref{prediction-problem-space}, we discuss the range of interesting problems in this dataset.


\subsection{Participants}

We recruited participants with backgrounds in data science, analytic, retail/ecommerce, and business to participant in this experiment. Overall, we had 22 users create reports and 26 users judge reports. A majority of the participants were hired from the freelance company Upwork.com so we could verify their qualifications. We paid users either a fixed price of \$50 USD to contribute or their hourly rate which ranged from \$15 to \$50 USD. The qualifications of other participants were verified by the survey every user completed before contributing.

\subsection{Role and Group assignment}

This experiment consisted of two primary roles: create and judge. Data scientists were responsible for creating predictive modeling reports and domain experts were responsible for judging them. 

After being assigned a role a user was not allowed to change roles. The one exception was for users could switch from create to judge after making 2 submissions. They could not switch back afterwards. Users who created and judged would not see their own submissions. 

With in the create role we assigned users to one of Group A, Group B, or Group C where the letter corresponds the interfaces described in Section \ref{exp_interfaces}. At the beginning of the experiment we randomly assigned users to group. During the experiment, we changed this strategy to make assignment that led to even group sizes. 

Each participant in the experiment was assigned an anonymous unique identifier. We used this identifier to track their experiment group, model submissions, judging votes, and survey responses. In the model creation interface, we collect timing data as described in \ref{timing_details}. Only this information is stored in our database for analysis.

\subsection{Prediction problem space}\label{prediction-problem-space}


We chose prediction problems that had potential to create business value, but allowed for a significant amount of variation in this value. This meant that the data science participants had to decide which problems they perceived as valuable and evaluating the models required domain experts to be involved.

In general, the problems fell under three categories, corresponding to the entities in the dataset they were defined over: users, orders, and departments. User problems predicted future user behaviour, order problems predicted information related to the products added to a user's cart during the rest of that particular order, and department problems predicted information related to the popularity of the produce department. 

Many of these problems present results that would be useful for completely different teams of people at Instacart or the brick-and-mortar grocery stores the dataset actually describes. Some involve supply-chain management, such as predicting how many orders will be placed in the produce department in the future. Others involve customer retention, such as predicting how many orders a user will make in the future. Still others involve recommendation (what products will a user buy or what item will be added to the cart next).

The problems spanned a wide range of predictability, varying from a low of absolutely no predictive accuracy, to highs of greater than 0.88 AUC for problems in each entity. This wide variance was useful from an experiment maker's perspective, as it made sure that the predictive accuracy of a problem played a role in both model creation and model judging. In other words, it was non-trivial, but also not too difficult, for the ``Data Scientist'' to build models with high predictive accuracy. Statistics about the predictive scores of these problems are presented in Tables \ref{table:pred_scores} and \ref{table:pred_std_scores}.

Note that each prediction problem defined a binary classification problem for simplicity and uniformity in the results. Future work may examine the relationship between choosing between binary classification, multi-class or multi-label classification, and regression.

\begin{table}[h]
\centering
\small
    \begin{tabular*}{\linewidth}{@{\extracolsep{\fill}}llllll@{}}
       \toprule
\textbf{\normalsize Entity} & \textbf{\normalsize Metric} & \textbf{\normalsize Mean} & \textbf{\normalsize Min} & \textbf{\normalsize Max} & \textbf{\normalsize Std} \\
 \midrule
orders & f1 & 0.645 & 0.000 & 0.921 & 0.255 \\
  & auc & 0.732 & 0.500 & 0.884 & 0.105 \\
  & accuracy & 0.728 & 0.512 & 0.863 & 0.074 \\
 \midrule
departments & f1 & 0.539 & 0.000 & 0.941 & 0.309 \\
  & auc & 0.704 & 0.500 & 1.000 & 0.158 \\
  & accuracy & 0.746 & 0.038 & 0.991 & 0.176 \\
 \midrule
users & f1 & 0.708 & 0.051 & 0.988 & 0.220 \\
  & auc & 0.721 & 0.500 & 0.925 & 0.108 \\
  & accuracy & 0.794 & 0.400 & 0.976 & 0.072 \\
  \bottomrule
    \end{tabular*}
    \caption{Statistics of various metrics of problems for the 3 different entities}
    \label{table:pred_scores}
\end{table}

\begin{table}[h]
\centering
\small
    \begin{tabular}{@{}lll@{}}
    \toprule
 \textbf{\normalsize Entity} & \textbf{\normalsize Metric} & \textbf{\normalsize Mean Std} \\
 \midrule
orders & f1 & 0.026 \\
  & auc & 0.006 \\
  & accuracy & 0.008 \\
 \midrule
departments & f1 & 0.122 \\
  & auc & 0.077 \\
  & accuracy & 0.082 \\
 \midrule
users & f1 & 0.022 \\
  & auc & 0.015 \\
  & accuracy & 0.012 \\
  \bottomrule
    \end{tabular}
    \caption{Mean standard deviations of various metrics of problems for the 3 different entities}
    \label{table:pred_std_scores}
\end{table}

\subsection{Machine learning method space }
We took advantage of Scikit-Learn \cite{scikit-learn}, an open-source machine learning framework, to quickly enumerate and build many different machine learning models. Algorithms were selected to optimize to varying degrees a combination of predictive accuracy, interpretability, and efficiency. We chose Random Forests, Support Vector Machines with a linear kernel, and 2-3 layer Multi-layer Perceptron (a neural network). We gave the user 2-3 degrees of freedom for each, opening up the hyperparameters that were generally regarded as sensitive to changes in the underlying data, yet disallowing selections that would drastically slow down runtime. To that effect, we restricted the SVM to a linear kernel so we could use a faster underlying implementation, and restricted the neural network to 3 layers. Refer to Table \ref{table:hyper_alg_choices} for details on which hyperparameters were used.

We performed a few processing steps on the feature matrices before machine learning. The first was a one-hot encoding of categorical features and imputation of missing values with 0 to make sure every value was numeric, and the second was a scaling of these values to within 0 and 1. Note that we performed no explicit dimensionality reduction or feature selection.

In the case of automatic modeling, we gave the user the median model as ranked by AUC.

\begin{table}[h!]
\centering
\small
    \begin{tabularx}{\linewidth}
       {@{}L{12em}AA@{}}
       \toprule
 \textbf{\normalsize Algorithm} & \textbf{\normalsize Hyperparameter} & \textbf{\normalsize Options} \\
 \midrule
Random Forest& n\_estimators & [10, 100, 500] \\
 & max\_depth & [3, 10, None] \\
  \midrule
Support-Vector Machine & C & [1, 0.1, 0.01] \\
 & loss & ['hinge', 'squared\_hinge'] \\
  \midrule
Multilayer Perceptron& solver & ['adam', 'sgd'] \\
 & activation & ['relu', 'tanh', 'logistic'] \\
  & hidden\_layer\_sizes & [[50, 50], [50, 100, 10], [50, 50, 20]] \\
  & alpha & [0.01, 0.001, 0.0001, 1e-05] \\
  \bottomrule
    \end{tabularx}
    \caption{Multiple modeling methods and hyperparameter choices per method.}
    \label{table:hyper_alg_choices}
\end{table}

\subsection{Precomputing models and their outputs}

To enable the interactivity of the experiment, we precomputed the results of any allowed combination prediction problem and hyperparameter selections in the interface. 

Building an app to allow arbitrary web users to compute on-demand presents numerous technical challenges, so we save that for a future date. Instead, we enumerated a list of 40 fully defined prediction problems on the Instacart grocery shopping dataset\cite{instacartdataset}, and 39 model/ hyperparameter options for each, for a total of 1584 end-to-end computations.

We used a modified version of the Deep Feature Synthesis algorithm \cite{kanter2015deep} to automatically build sets of 39, 34, and 11 features for problems defined per-order, per-user, and per-department respectively. Note that multiple sets of features were computed for each department corresponding to unique points in time.

With labels and features, we trained out each of the 39 model \& hyperparameter combinations on each prediction problem, and scored each for AUC, F1, and accuracy using 5-fold cross-validation.

Lastly, we generated feature importances for each model using random forests and SVMs. Feature importances are built into tree-based machine learning algorithms, and so were easily accessible for random forest models. We used a linear kernel and the l2 norm for regularization in the SVM, so feature importances are simply the coefficients of the hyperplane. Feature importances are more difficult to obtain for MLP, and in general for all neural network models, so we left them out. This mimics what a data scientist would submit as preliminary results in the real world.

\section{Results}\label{results}
In this section, we present the results of deploying Prediction Factory on the Instacart dataset.

\subsection{Model Creation}
We had a total of 22 users participate in creating models. The total number of users and submissions for each experiment groups is given in Table \ref{user_counts}. 

\begin{table}[!h]
\centering
\small
    \begin{tabular}
        {@{}lll@{}}
        \toprule
         & \textbf{\normalsize \# of Participant} & \textbf{\normalsize \# of Submissions}  \\
        \midrule
        Group A & 6 & 27 \\
        Group B & 9 & 37 \\
        Group C & 7 & 30 \\
        All & 22 & 94 \\
        \bottomrule
    \end{tabular}
    \caption{Number of participants and submission for each group.}\label{user_counts}
\end{table}

We recorded the time user spent in each of the 3 key tasks. Table \ref{timing} shows the mean total time per submission in each group as well as the percentage of the total each task contributed.

\begin{table}[!htb]
\centering
\small
    \begin{tabular*}{\linewidth}{@{\extracolsep{\fill}}lllll}
        \toprule
         & \textbf{\normalsize Mean time (sec)} & \textbf{\normalsize \% Specifying} & \textbf{\normalsize \% Learning}  & \textbf{\normalsize \% Writing} \\
        \midrule
        Group A & 259 & 11.6 & 33.3 & 55.1 \\
        Group B & 285 & 13.6 & 14.3 & 72.1 \\
        Group C & 318 & 19.3 & 23.0 & 57.7 \\
        All & 288 & 15.1 & 22.2 & 62.7 \\
        \bottomrule
    \end{tabular*}
    \caption{Mean total time per submission and mean percentage of time spent per task broken down by group.}\label{timing}
\end{table}

\subsection{Model Judging}

We had 26 users participate in judging 296 pairs of models. The results of judging where the models were from different groups are given in Table \ref{judging_results}.

\begin{table}[!htb]
\centering
\small
    \begin{tabular*}{\linewidth}{@{\extracolsep{\fill}}llll}
       \toprule
        & \textbf{\normalsize A vs B} & \textbf{\normalsize A vs C} & \textbf{\normalsize B vs C} \\
        \midrule
        Win \% & 42.5 & 39.7  & 55.0 \\
        Total votes & 80 & 75 & 60 \\
        \bottomrule
    \end{tabular*}
    \caption{Head-to-head judging results for the three possible pairings of groups. Win percentage is for first group in pair.}\label{judging_results}
\end{table}


\subsection{Ranking individual users}
In addition to group-wide comparisons, we compare how individual users perform in head to head match ups using the Elo rating system \cite{elo}. To do this, we consider each vote as a game with a zero-sum binary outcome. Elo ratings are then derived from the probability that a user will win the current game, considering his or her past history of winning.

We computed Elo scores with the standard calculation method used by FIDE (World Chess Federation). We set a minimum score of 100 and an initial score of 500. Various statistics are presented in Table \ref{table:elo_stats}.

\begin{table}[h]
\centering
\small
\begin{tabularx}{\linewidth}{@{}AAAAAAl@{}}
\toprule
\textbf{\normalsize Group} & \textbf{\normalsize Min} & \textbf{\normalsize 25th} & \textbf{\normalsize Mean} & \textbf{\normalsize 75th} & \textbf{\normalsize Max} & \textbf{\normalsize STD}\\
& & \textbf{\normalsize percentile} & & \textbf{\normalsize percentile} & & \\
\midrule 
Overall & 100 & 197 & 667 & 930 & 2072 & 613\\
A & 100 & 200 & 362 & 521 & 694 & 211\\
B & 131 & 199 & 745 & 1009 & 2072 & 621\\
C & 100 & 199 & 828 & 1424 & 1973 & 730\\
\bottomrule
\end{tabularx}
\caption{Overall and per-group statistics on users' Elo ratings}
\label{table:elo_stats}
\end{table}

\subsection{Usage of optional automation}
Users in Group C were not required to use the automated modeling functionality. Of the 7 users in Group C, all but one (85.7\%) used it for at least one of their submissions. Across all submissions by users in Group C, 60\% used automated modeling. The head-to-head results of Group C automated versus Groups A and B are in table \ref{automodel_c}.

\begin{table}[h]
\centering
\small
    \begin{tabular*}{\linewidth}{@{\extracolsep{\fill}}lll}
       \toprule
        & \textbf{\normalsize A vs C Auto model} & \textbf{\normalsize B vs C Auto model} \\
        \midrule
        Win \% & 38.8 & 50.0  \\
        Total Votes & 49 & 38 \\
        \bottomrule
    \end{tabular*}
    \caption{Automodel C vs Groups A and B. Win percentage given for first group in pair.}\label{automodel_c}
\end{table}

\section{Key findings}\label{discussion}
Automated prediction engineering, feature engineering, and machine learning has the potential to profoundly change the experience of building predictive models. The results of the experiment demonstrate significant differences in how data scientists in each group used the report creation interfaces, as well the the funding outcomes themselves. In this section we discuss the key findings from this experiment 

\textbf{Automation changed how data scientists spent their time.} Users in Group B who had full automation spent about the same time overall as their peers in group A, but 31\% more time than group A on writing their report. Importantly, while the users saved time using the automated machine learning they decided to reallocate that time to writing their reports.

Users in Group C spent a similar percentage of their time writing, but spent more time across the 3 key tasks than any other group. This is likely because in 40\% of submissions, these users manually selected machine learning parameters.

\textbf{Automated machine learning is preferred, but optimal performance isn't required.} Unlike Groups A and B, users in Group C did not have to use automation for machine learning. Still, in a majority of cases, they decided to despite the fact that the automated solution only provided the median performing model. Even more, the submissions from Group C that used automated machine learning performed slightly better in the judging than the ones that didn't.

This demonstrates machine learning hyperparameter optimization is likely premature at the beginning stages of a predictive modeling project. Instead spending more time communicating and explaining results has a bigger impact on funding success. 

\textbf{Automation improved funding outcomes.} On average, predictive solutions generated by the group using full automation were funded 35\% percent more when compared against the group that used baseline automation to create solutions, while the solutions from group that had optional automation were funded 58.6\% more often compared to the baseline. Solutions generated using full automation and optional automation were funded about equally when put head-to-head.

This indicates that automation can play an important role in helping data scientists increase the business value of their work. 

\textbf{Have we reached a critical amount of automation?} The key difference between users in Group A vs Groups B and C was the ability to revisit the problem specification parameters after seeing results from modeling. While Group C was allowed to manually find a optimal machine model, we see searching for it did not improve outcomes. 

The results of the Elo analysis demonstrate that automation led users in Groups B and C to out rank their peers in Group A. On average, a user in Group A ranked 383 and 466 points below their peers in Groups B and C, respectively. 

Ultimately, users who had access to a feedback loop around problem definition and modeling, that is-- who were able to redefine the problem they were solving after seeing its feasibility-- performed significantly better. Moreover, access to machine learning parameters had virtually no effect on their scores.

\section{Conclusion}\label{conclusion}
Data science remains a complex and unstructured field - one that many businesses struggle to incorporate into their work. A lack of shared understanding makes it difficult for experts from different domains to work together, even though teamwork would improve results. Prediction Factory is a system for data scientists and domain experts to collaborate using predictive modeling automation. 

Enabled by key data science automation technologies, Prediction Factory encourages collaboration via an interactive loop: data scientists translate business goals into predictive modeling tasks and business experts provide feedback on the results to incorporate into future work. The outcome is a curated set of impactful predictive models.
\input{discussion}
\begin{table*}[!ht]
\small
\begin{tabularx}{\linewidth}{@{}lL{0.12\linewidth}AL{0.16\linewidth}L{0.16\linewidth}@{}}
\toprule
\textbf{\normalsize Entity} & \textbf{\normalsize Sentence} & \textbf{\normalsize Parameter $\mathtt{XX}$} & \textbf{\normalsize Parameter $\mathtt{YY}$} & \textbf{\normalsize Parameter $\mathtt{ZZ}$}\\
\midrule 
Orders & Predict if $\mathtt{XX}$ $\mathtt{YY}$.\newline When should we make this prediction? $\mathtt{ZZ}$ & \vspace{-0.7\baselineskip}\begin{itemize}[noitemsep,leftmargin=*,topsep=0pt,parsep=0pt,partopsep=0pt,label=--]
    \item produce is the most common department the user buys from
    \item the user
adds at least 3 more items
    \item the user adds a product from the produce
department
    \item the user adds more than 3 unique departments
\vspace{-\baselineskip}\end{itemize} & 

\vspace{-0.7\baselineskip}\begin{itemize}[noitemsep,leftmargin=*,topsep=0pt,parsep=0pt,partopsep=0pt,label=--]
    \item before checking out
    \item in the next 3 cart adds
\vspace{-\baselineskip}\end{itemize}
 & 
 
\vspace{-0.7\baselineskip}\begin{itemize}[noitemsep,leftmargin=*,topsep=0pt,parsep=0pt,partopsep=0pt,label=--]
     \item at the start of the order.
     \item after adding 3 items to the cart.
\vspace{-\baselineskip}\end{itemize}
 
 \\
\midrule
Departments & Predict if $\mathtt{XX}$ $\mathtt{YY}$.\newline How much training data? $\mathtt{ZZ}$ &

\vspace{-0.7\baselineskip}\begin{itemize}[noitemsep,leftmargin=*,topsep=0pt,parsep=0pt,partopsep=0pt,label=--]
    \item more than 1000 users buy from the produce department
    \item more than
1000 orders are placed in the produce department
    \item more than 100 orders
are placed in the produce department
    \item more than 100 users buy from
the produce department
\vspace{-\baselineskip}\end{itemize} & 

\vspace{-0.7\baselineskip}\begin{itemize}[noitemsep,leftmargin=*,topsep=0pt,parsep=0pt,partopsep=0pt,label=--]
    \item in the next 7~days
    \item in the next 14~days
\vspace{-\baselineskip}\end{itemize}
 &
 
 \vspace{-0.7\baselineskip}\begin{itemize}[noitemsep,leftmargin=*,topsep=0pt,parsep=0pt,partopsep=0pt,label=--]
     \item At least 7~days of training data.
     \item At least 30~days of training
 \vspace{-\baselineskip}\end{itemize}
\\
\midrule
Users & Predict if a user will make $\mathtt{XX}$ of $\mathtt{YY}$ items in $\mathtt{ZZ}$~days. & 
\vspace{-0.7\baselineskip}\begin{itemize}[noitemsep,leftmargin=*,topsep=0pt,parsep=0pt,partopsep=0pt,label=--]
    \item more than 3 orders
    \item an order
    \item more than 1 orders  
\vspace{-\baselineskip}\end{itemize}

 & 
\vspace{-0.7\baselineskip}\begin{itemize}[noitemsep,leftmargin=*,topsep=0pt,parsep=0pt,partopsep=0pt,label=--]
    \item any number of
    \item more than 5
    \item more than 10
    \item more than 15
\vspace{-\baselineskip}\end{itemize}
& 
\vspace{-0.7\baselineskip}\begin{itemize}[noitemsep,leftmargin=*,topsep=0pt,parsep=0pt,partopsep=0pt,label=--]
    \item any number of
    \item the next 30
\vspace{-\baselineskip}\end{itemize}
\\
\bottomrule
\end{tabularx}

    \caption{Prediction problem definitions in natural language. Problems are presented as sentences with variable sections representing parameter settings. Each sentence contains three variables, $\mathtt{XX}$, $\mathtt{YY}$, and $\mathtt{ZZ}$, and the possible options for each are displayed in the three columns to the right of the sentence.}
    \label{table:pred_defns}
\end{table*}

\bibliographystyle{ACM-Reference-Format}
\bibliography{main}

\end{document}